# Cross-Entropy Optimization for Hyperparameter Optimization in Stochastic Gradient-based Approaches to Train Deep Neural Networks

Kevin Li, Fulu Li

Contact: fulu@alum.mit.edu

**Abstract**

In this paper, we present a cross-entropy optimization method for hyperparameter optimization in stochastic gradient-based approaches to train deep neural networks. The value of a hyperparameter of a learning algorithm often has great impact on the performance of a model such as the convergence speed, the generalization performance metrics, etc. While in some cases the hyperparameters of a learning algorithm can be part of learning parameters, in other scenarios the hyperparameters of a stochastic optimization algorithm such as Adam [5] and its variants are either fixed as a constant or are kept changing in a monotonic way over time. We give an in-depth analysis of the presented method in the framework of expectation maximization (EM). The presented algorithm of cross-entropy optimization for hyperparameter optimization of a learning algorithm (CEHPO) can be equally applicable to other areas of optimization problems in deep learning. We hope that the presented methods can provide different perspectives and offer some insights for optimization problems in different areas of machine learning and beyond.

**1. Introduction**

In machine learning, the most fundamental components of the system are the learning algorithms. A hyperparameter of a learning algorithm is often used to control the learning process, whose value may have great impact on the performance of the learning algorithm such as the convergence speed and other performance metrics such as the value of a pre-defined loss function for a given data set. Another aspect is the estimation of the generalization performance of the model, which is often accomplished by using cross-validation techniques with a totally different data set. The goal is to choose a set of values for the hyperparameters of a learning algorithm to maximize the generalization performance. In some cases, some hyperparameters of a learning algorithm can be included as part of learning parameters with multiple independent learning processes with different sets of hyperparameters. While in some other scenarios, it may not be applicable for some hyperparameter of a learning algorithm to be part of learning parameters. Therefore, it is a common practice to apply some hyperparameter optimization techniques to choose optimal or sub-optimal hyperparameters for a given learning algorithm.

For example, the stochastic optimization method of Adam in [5] uses the hyper-parameters $\beta_1, \beta_2 \in [0, 1)$ to control the exponential decay rates of these moving averages of the first moment (the mean) and the second raw moment (uncentered variance) of the gradients respectively. By choosing proper values of the hyperparameters of $\beta_1, \beta_2$, the basic idea is to assign smaller weights to the gradients too far in the past [5,7]. As discussed in [5,7], in some cases both $\beta_1, \beta_2$ are held constant and in other cases $\beta_1$ might be kept decreasing over time. In the experiments in [7], a grid search method is used in choosing the hyperparameters of $\alpha$ and $\beta_2$.

Nowadays, stochastic gradient-based optimization methods are the key pillars to train deep neural networks and a hyper parameter is a parameter whose value is used to control the learning process during the training of deep neural networks. As discussed in [11], traditional stochastic gradient descent (SGD) algorithm for training deep neural networks requires careful tuning of the learning rate and momentum parameters. Momentum-based approaches accumulate gradients over time to maintain the momentum for the optimizer and to avoid being stuck in some local minima. In [5], a widely used approach named Adam was presented to adapt the learning rate for each parameter of the deep neural networks based on first and second moments of the gradients. In [7], Reddi et al discussed some of the convergence issues of Adam and pointed out that some of the convergence problems of approaches that employed exponential moving average of the past gradients are due to the fact that large gradients die out quickly with the effects of exponential moving average, which may lead to poor convergence.

In this paper, we present a novel framework for the optimization of hyperparameters in stochastic gradient-based methods to train deep neural networks such as Adam and its variants. Our key contributions are: We apply cross-entropy (CE) optimization techniques for the hyperparameter optimization in Adam and its variants, which is widely used to train deep neural networks. The presented framework can be applied for other optimization problems in deep learning and beyond. We hope that the presentation of cross-entropy optimization for hyperparameters optimization of learning algorithms can provide different perspectives and offer some insights for optimization problems in different areas of machine learning and beyond.

The rest of the paper is organized as follows: We discuss the related work in Section 2. We present cross-entropy optimization framework for hyperparameter optimization in Adam in Section 3. We give an in-depth analysis of the presented method in the framework on expectation maximization (EM) in Section 4. The conclusions and future directions are given in Section 5.

## 2. The Related Work

Existing hyperparameter optimization methods include: (a) grid search, i.e., essentially an exhaustive search over a given hyperparameter space; (b) random search, which can explore much more potential values than grid search for continuous hyperparameters; (c) Bayesian optimization, which employs a probabilistic model between hyperparameter values and objectives evaluated on a given data set and it has shown to outperform both grid search and random search with fewer evaluations; (d) gradient-based optimization, which computes the gradients with respect to hyperparameters when applicable for some learning algorithms and then optimizes the hyperparameters based on gradient decent philosophy; (e) evolutionary optimization, which employs an evolutionary algorithm to find the best possible hyperparameters for a given learning algorithm; (f) population-based methods, which updates the hyperparameters as well as the weights of neural networks during the training of the models with multiple independent learning processes with different hyperparameteres of the learning algorithm; (g) early stopping-based approach, which starts as a random search over the hyperparameters space and successively prunes low-performing ones until there is only one model remaining.

Notably, the basic idea of cross entropy (CE) method [6, 9, 10] is to translate the deterministic optimization problem into a corresponding stochastic one and then use rare event simulation techniques to find the optimal solution. In [6], Li et al presented a random tree optimization approach for the construction of the most parsimonious phylogenetic trees based on CE method, where an iterative optimization process is applied to reduce the parsimony score of a tree with the principle of cross entropy optimization. As discussed in [6, 9, 10], cross entropy (CE) method differs from other well-known random search algorithms for global optimization such as simulated annealing [1,2,8], tabu search [3] and genetic algorithms [4], which are local search heuristics and employ the notion of local neighborhood structures. Cross entropy (CE) method employs multi-extremal optimization process based on Kullback-Leibler cross-entropy, importance sampling, etc. Therefore, cross entropy (CE) method represents a *global* random search procedure rather than a *local* one.

Our work differs from existing hyperparameter optimization methods in that we use cross-entropy optimization with a probability distribution over hyperparameter space, where some rare event simulation techniques are used to find the optimal values of the hyperparameters for a given learning algorithm. The cross-entropy operation is achieved by virtue of probability update after the performance evaluation after each round of randomly-generated hyperparameter samples based on probability distributions, where a small portion of top performers are given higher probabilities for the next round. The iterative optimization process ends when the convergence conditions are achieved or a pre-defined early-stop criteria is met.

## 3. Hyperparameters Optimization for Adam

We have some basic notations for the cross entropy optimization of hyperparameters in Adam and its variants [5,7] as follows: Let $\beta$ be a hyperparameter for a given learning algorithm. Let $F(\beta)$ stand for the performance metric for a given value of $\beta$, a given learning algorithm and a given data set for a given machine learning problem. Let $n_d$ be the number of data sets for a given machine learning problem. Let $n_{ml}$ be the number of machine learning problems. We also assume that the proper value range of $\beta$ is given as $[a, b]$, where $b > a$ and $a, b \in \mathfrak{R}$. Let $\gamma_t$ stand for the benchmark value of $F(\beta)$ for the $t^{th}$ round. In the case of convergence time, we have its definition as follows:

$$\gamma_t = \min\{f : P_{\beta_{t-1}}(F(\beta) \leq f) \geq \rho\} \quad (1)$$

where the purpose of the hyperparameter optimization is minimize the convergence time of the given learning model, $\rho$ normally takes a value of 0.01 so that the event of obtaining high performance is not too rare, the value of $\beta$ is randomly chosen based on the probability distribution of the $\beta$ values, $\beta_{t-1}$ represents the set of randomly generated values of $\beta$ in the $(t-1)^{th}$ round. Essentially, $\gamma_t$ is the top $\rho$-quantile of the performers of the randomly generated $\beta$ values in the $t^{th}$ round.

In the case of generalization performance, we have the definition of $\gamma_t$ as follows:

$$\gamma_t = \max\{f : P_{\beta_{t-1}}(F(\beta) \leq f) \geq \rho\} \quad (2)$$

where the purpose of the hyperparameter optimization is to maximize the generalization performance of the given learning model.

As discussed in [6,9,10], there are several ways to set the termination conditions. Normally, if for some $t \geq l$, say $l = 5$, and we have

$$\gamma_t = \gamma_{t-1} = \ldots = \gamma_{t-l} \quad (3)$$

then we stop the hyperparameter optimization process.

Let $M$ stand for the number of sample values of $\beta$ in a given round, let $\beta_i$ denote the $i^{th}$ randomly generated sample value of $\beta$, $H_{\{\}}$ is an indicator function, $q_i$ is the probability that the given $i^{th}$ value of $\beta_i$ is being chosen and $q_i$ is initialized as zeros at the beginning. We assume uniform distribution of the values of $\beta$ in a given range of $[a, b]$, where $b > a$ and $a, b \in \mathfrak{R}$. Randomly choosing a value of $\beta$ in a given range of $[a, b]$ with uniform distribution is like generating a random number in the range of $[a, b]$ with uniform distribution. The updated value of $q_i$ can be estimated as:

$$q_i^e = \frac{H_{\{F(\beta_i) \leq \gamma\}}}{\sum_{k=1}^{M} H_{\{F(\beta_k) \leq \gamma\}}} \quad (4)$$

While there are solid theoretical justifications for Equation (4), we refer interested readers to [9,10] for the details.

In order to have a smoothed update procedure, normally we have

$$q_i^t = c \times q_i^e + (1 - c) \times q_i^{t-1} \quad (5)$$

where empirically a value of $c$ between 0.4 and 0.9, i.e., $0.4 \leq c \leq 0.9$, gives the best results [9,10], $q_i^{t-1}$ is the value of $q_i$ in the previous round, $q_i^e$ is the estimated value of $q_i$ based on the performance in the previous round according to Equation (4), $q_i^t$ stands for the value of $q_i$ in the current round.

Let $B$ be the set of $\beta$ values whose performance are in the top $\rho$-quantile of the performers of the randomly generated $\beta$ values in a given round and we have

$$B = \{\beta_i | H_{\{F(\beta_i) \leq \gamma\}}\} \quad (6)$$

A normalized $q_i^t$ for the top $\rho$-quantile of the performers is given as follows:

$$q_i^n = \frac{q_i^t}{\sum_{\beta_i \in B} q_i^t} \quad (7)$$

Let $s$ be the factor of favorability, i.e., $s = 10$, towards those top performers in a given round and let $N_s$ be the total number of samples chosen from those top performers and we have

$$N_s = s \times M \times \rho \quad (8)$$

For the next round, we have $N_s$ samples that are randomly chosen from those $|B|$ number of $\beta$ values, where $|B|$ indicates the number of elements in the value set of $B$. The rest of the samples, i.e., $(M - N_s)$ samples, are randomly chosen from the range of $[a, b]$ with uniform distribution, which can be accomplished by generating $(M - N_s)$ random numbers in the range of $[a, b]$ with uniform distribution.

In summary, we have the algorithm of cross entropy optimization for the hyperparameters optimization (CEHPO) in Adam as follows:

For a given data set and a given machine learning problem, we have the CEHPO algorithm as follows:

1. Set $t = 1$ and initialize $q_i$ as zeros.
2. Generating $M$ random numbers in the range of $[a, b]$ with uniform distribution for the values of $\beta$ if it is the first time.
3. Generating $(M - N_s)$ random numbers in the range of $[a, b]$ with uniform distribution for the values of $\beta$ if it is not the first time.
4. $N_s$ is determined by Equation (8).
5. Calculate $\gamma_t$ according to either Equation (1) or Equation (2), depending on the optimization scenarios.
6. Update $q_i$ according to Equation (4) and Equation (5).
7. Update normalized $q_i$ according to Equation (7).
8. If for some $t \geq l$, say $l = 5$, such that $\gamma_t = \gamma_{t-1} = \ldots = \gamma_{t-l}$, then stop, otherwise, reiterate from Step 3.

**Figure 1**: the CEHPO Algorithm 1 for a given data set and a machine learning problem.

For multiple data sets and multiple machine learning problems, we have the CEHPO algorithm as follows:

1. for $d_i$ in $[d_1 : d_{n_d}]$
2. // $n_d$ is the number of data sets
3. {
4.   for $m_j$ in $[m_1 : m_{n_{ml}}]$
5.   // $n_{ml}$ is the number of machine learning problems
6.   {
7.     Execute CEHPO Algorithm 1 described in Figure 1.
8.     Record the best choice of $\beta_{d_i, m_j}$.
9.   }
10. }
11. Choose $\beta^*$ based on Equation (9).
12. Output $\beta^*$.

**Figure 2**: the CEHPO Algorithm 2 for multiple data sets and multiple machine learning problems.

Let $\beta_{d_i, m_j}$ stand for the best choice of $\beta$ value for a given data set of $d_i$ and a given machine learning problem of $m_j$, the best choice of $\beta^*$ among different data sets and different machine learning problems can be computed as follows:

$$\beta^* = \min_x \sum_{i=1}^{n_d} \sum_{j=1}^{n_{ml}} (\beta_{d_i, m_j} - x)^2 \quad (9)$$

where $x$ is one of the $\beta_{d_i, m_j}$ values. Essentially, we are choosing the one that has the least distance to all the other best choices of $\beta$ values as the final output of $\beta^*$.

In the following, we discuss on how to deal with the scenario of decreased $\beta_1$ over time in Adam with cross entropy hyperparameter optimization (CEHPO). We can replace the original scalar value of $\beta$ with a limited sequence of $\beta$ values in decreasing fashion, each of which can be mapped to a certain period/epochs during the learning process. Essentially, the optimal values of $\beta$ is a sequence of values in decreasing order. The other part of the logic for CEHPO algorithm can be equally applicable to this scenario.

According to [5,7], the main difference between AMSGRAD [7] and Adam [5] is that AMSGRAD keeps the status of the maximum of all $v_t$ until the present time step and uses the maximum value to normalize the running average of the gradient instead of $v_t$ in Adam. This leads to non-increasing step size for AMSGRAD in order to avoid the pitfalls of Adam and others.

**4. Analysis in the Framework of EM**

Following [6], we give an in-depth analysis on CEHPO algorithm in the framework of expectation maximization (EM) [12] in this section, which could shed some light on the theoretical justifications of some of the operations in the process of cross entropy optimization.

Let $D$ indicate the values of the observed variables, e.g., the data sets in the learning experiments, and let $W$ denote the hidden data, e.g., the weights in the neural networks. Let $P$ be the probability mass function of the complete data with some hyperparameters given by $\beta$ and others such as $\theta$, etc. For the sake of simplicity, we only consider hyperparameters of $\beta$ and omit other parameters such as $\theta$, etc. in the following discussions. Therefore, we have $P(D, W | \beta)$ as the complete data likelihood, which can be thought of as a function of $\beta$.

By using the Bayes's rule and the law of total probability, the conditional probability of the hidden data given the observed data and $\beta$ can be expressed as:

$$P(W|D,\beta) = \frac{P(D,W|\beta)}{P(D|\beta)}$$

$$= \frac{P(D|W,\beta)P(W|\beta)}{\sum_{\hat{W}} P(D|\hat{W},\beta)P(\hat{W}|\beta)} \qquad (10)$$

where $\hat{W}$ indicates the estimated weights of the neural networks.

The goal is to estimate the values of $\beta$. The E-step is given by:
$$E: \quad Q(\beta) = E_W[\log(P(D,W|\beta))] \qquad (11)$$
where $Q(\beta)$ is the expected value of the log-likelihood of the complete data.

The expected value of log-likelihood in Equation (11) can be further expressed as:
$$Q(\beta) = \sum_W P(W|D,\beta) \times \log(P(D,W|\beta)) \qquad (12)$$

In the CEHPO algorithm, $\beta$ evolves from $\beta_0$ to $\beta_1, \beta_2, \ldots, \beta_{t-1}, \beta_t$ based on the improved $\beta$ samples according to Equation (4), Equation (5), Equation (6) and Equation (7) for better performance of the given learning algorithm for the given machine learning problem. The update process of $q_i$s according to Equation (4) and Equation (5) minimizes the cross entropy based on the Kullback-Leibler cross entropy principle [9, 10] and importance sampling philosophy.

The M-step is given by:
$$M: \quad \beta_t = \arg\max_\beta Q(\beta) \qquad (13)$$

where $\beta_t$ is the value of $\beta$ that maximizes (M-step) the conditional expectation (E-step, e.g., Equation (11)) of the complete data log-likelihood given observed data set under the previous parameter value of $\beta_{t-1}$. Notably, for each set of weights for the neural network, there is a likelihood value for $\beta$. We can thus calculate an expected value of the likelihood, which depends on the previously assumed value of $\beta$ as it influenced the probabilities of the weights of the neural network, e.g, $W$.

As discussed in [6], for traditional EM algorithm, it can be shown that an EM iteration does not decrease the observed data likelihood function. However, there is no guarantee that the data set converges to a maximum likelihood estimator. In other words, EM is a local search algorithm. In the CEHPO algorithm based on cross entropy optimization method, a multi-extremal search is employed due to the adoption of Kullback-Leibler cross-entropy [9,10] and importance sampling techniques. It can be shown in [9,10] that with high probability the observed data likelihood function increases and eventually converges to one so long as the number of samples of *β* value and the data set of *D* are large enough and the number of iterations is large enough.

We also need to emphasize that the following properties hold with high probability for CEHPO algorithm based on cross entropy optimization method:

$$P(D|\beta_i) = \sum_W P(D, W|\beta_i) \leq P(D|\beta_{i+t})$$
$$= \sum_W P(D, W|\beta_{i+t}) \quad (14)$$

When both *t* and the number of samples of weights of the neural network, e.g., |*W*|, go to infinity, the data likelihood function converges. This property of cross entropy optimization method is proved in [9,10] based on the stochastic nature of Kullback-Leibler cross entropy process. We refer interested readers to [9, 10] for details.

## 5. Conclusion and Future Directions

In this paper, we present a cross-entropy optimization method for hyperparameter optimization in stochastic gradient-based approaches to train deep neural networks. The value of a hyperparameter of a learning algorithm often has great impact on the performance of a model such as the convergence speed, the generalization performance metrics, etc. While in some cases the hyperparameters of a learning algorithm can be part of learning parameters, in other scenarios the hyperparameters of a stochastic optimization algorithm such as Adam [5] and its variants are either fixed as a constant or are kept changing in a monotonic way over time. We present detailed operation flows of the cross entropy optimization in the CEHPO algorithm (Figure 1 and Figure 2) in a variety of application scenarios. We also give an in-depth analysis of the presented method in the framework of expectation maximization (EM) to shed some light on some of the theoretical justifications of the cross entropy optimization philosophy, which uses some of the rare event simulation techniques to find the best possible solution. The presented algorithm of cross-entropy optimization for hyperparameter optimization of a learning algorithm (CEHPO) can be equally applicable to other areas of optimization problems in deep learning. We hope that the presented methods can provide different perspectives and offer some insights for optimization problems in different areas of machine learning and beyond.

We will conduct extensive experiments with the presented CEHPO algorithm with a variety of data sets over a variety of machine learning problems in a variety of hyperparameter optimization scenarios as our future directions.